# Adaptive actuation of magnetic soft robots using deep reinforcement learning

Jianpeng Yao[1,2], Quanliang Cao[1,2,*], Yuwei Ju[1,2], Yuxuan Sun[1,2], Ruiqi Liu[1,2], Xiaotao Han[1,2,*], Liang Li[1,2]

1. Wuhan National High Magnetic Field Center, Huazhong University of Science and Technology; Wuhan 430074, PR China;
2. State Key Laboratory of Advanced Electromagnetic Engineering and Technology, Huazhong University of Science and Technology; Wuhan 430074, PR China.
* Corresponding authors. Email: quanliangcao@hust.edu.cn (Q.C.); xthan@hust.edu.cn (X.H.)

**Abstract**
Magnetic soft robots have attracted growing interest due to their unique advantages in terms of untethered actuation and excellent controllability. However, finding the required magnetization patterns or magnetic fields to achieve the desired functions of these robots is quite challenging in many cases. No unified framework for design has been proposed yet, and existing methods mainly rely on manual heuristics, which are hard to satisfy the high complexity level of the desired robotic motion. Here, we develop an intelligent method to solve the related inverse-design problems, implemented by introducing a novel simulation platform for magnetic soft robots based on Cosserat rod models and a deep reinforcement learning framework based on TD3. We demonstrate that magnetic soft robots with different magnetization patterns can learn to move without human guidance in simulations, and effective magnetic fields can be autonomously generated that can then be applied directly to real magnetic soft robots in an open-loop way.

**MAIN TEXT**
Soft robotics has recently become a hot topic for both practice and research in the field of robotics due to the distinct advantages of soft robots over traditional rigid robots, such as high deformability, dexterity, and robustness (*1, 2*). With the development of responsive soft materials, various kinds of soft robots have been invented. The relevant external stimuli include but are not limited to pressure (*3, 4*), heat (*5, 6*), light (*7, 8*), electric field (*9, 10*), and magnetic field (*11, 12*). Among them, magnetic soft robots are considered one of the most promising soft robots in biomedical applications because of several unique advantages of magnetic actuation, such as high controllability and safe, non-contact features (*13, 14*). Therefore, magnetic soft robotics has been extensively studied over the past five years, which involves multiple aspects, including magnetization methods (*12, 15–19*), actuation methods (*11, 20–22*), material design (*23–25*), and application development (*26–29*). The progress in these areas has been well-reviewed (*14, 30–32*). However, with the increasing complexity of application scenarios and functions of magnetic soft robots, it is challenging to design the required magnetic torques (including magnetization pattern and magnetic fields) acting on these robots according to specific manipulation requirements, which is a highly nonlinear inverse problem. Existing related research mainly focuses on manual-heuristic approaches (*14, 30*), which leads to a mismatch between design concepts and advanced manufacturing technologies.

In this work, we proposed an intelligent design method based on reinforcement learning (RL) to solve the above problem. RL algorithms are well known for their capabilities to solve problems without explicit human guidance (*33, 34*). Designers abstract control objectives to a reward function, and then agents, the learner and decision-maker of RL, will learn to maximize reward through interaction with environments. The state-of-the-art RL algorithms are able to solve real-life problems, and several thrilling applications have been published in other areas (*35*, *36*). In our paper, we choose TD3 (*37*), a modern deterministic model-free algorithm, as our base RL algorithm.

When implementing RL algorithms to robots and other real equipment, an efficient simulator is often needed for training. Former researchers of MSRs mainly choose an ABAQUS subroutine written by Zhao (*12*, *38*) to simulate deformations of the robots. However, finite-element methods (FEMs) are not computationally efficient. It is suitable to conduct mechanical analysis but not an excellent choice to serve as a simulation environment for RL. Models from the computer graphics community like discrete differential geometry (*39–41*) and Cosserat rod (*42–45*) give rise to new simulation methods for soft robots, which are quicker to compute with fair precision and provide possibilities for the implementation of modern AI techniques like RL (*46*). But previous models seldomly take specific external actuation into account, limiting their applicability with real soft robots. Our work combines magnetic torques and a novel dissipation model with the original Cosserat rod model, enabling the model to simulate MSRs with 2D deformation and movement.

With the simulation platform, the kinematics of soft robots can be easily extracted, such as position, orientation, and velocity. Using heuristic analysis based on our understandings of MSRs, we abstract useful information as features and input them to TD3 as state variables. We choose an elementary and straightforward reward function so that RL agents can focus on moving in a predefined direction. RL agents control the external magnetic fields for actuation, which then control the movement of the robots. During training, RL agents interact with environments in simulations and, from past experiences, learn to control robots to move forward. After training, we obtain some stable control policies, which are used to generate magnetic fields. We present a few RL cases given different magnetization patterns and different field amplitude limitations to show the generality of this method. The results demonstrate that RL agents can adapt to different conditions automatically, and the fields generated can validly actuate MSRs to move forward, both in simulations and experiments.

Our work shows that RL can be used to help design magnetic parameters of MSRs. Using RL, researchers can focus on the goals robots are ordered to achieve rather than the complex mechanics of nonlinear motion. Given an objective, soft robots can learn to fulfill that without any human design. Besides, the learning results differ according to different conditions, which indicates that RL agents can adaptively learn to utilize various useful features, which is highly promising considering the flexibilities and robustness of soft robots themselves.

**Numerical model for predicting dynamic behaviors of MSRs**

Cosserat rod model owns the benefits of covering common types of deformation like tension, bending, shear, and torsion (*42*). It abstracts slender structures to a centerline curve and the corresponding cross-sections. Elements of an elastic rod with an unstressed length of $l$ can be labeled with arc-length parameter $s \in [0,l]$ and time $t$. At each point, a vector $\mathbf{r}(s,t)$ can be used to describe the position under the Eulerian coordinates, and a triad of orthogonal unit vectors $\mathbf{d}^i(s,t)$, ($i$=1,2,3) can be used to describe the orientation of cross-sections. For convenience, $\mathbf{d}^3$ is chosen to be perpendicular to the cross-section, and $\mathbf{d}^1$ and $\mathbf{d}^2$ lie in the cross-section plane. $\mathbf{d}^3$ points to different directions from the centerline tangent $\partial_s \mathbf{r}$ due to shear. Under numerical settings, we discretize a rod into nodes and elements. $\mathbf{r}(s,t)$ is of nodal quantities, used to describe the coordinates of nodes; $\mathbf{d}^i(s,t)$ belongs to elemental quantities, used to describe the orientation of elemental cross-sections, as is shown in Fig. 1A.

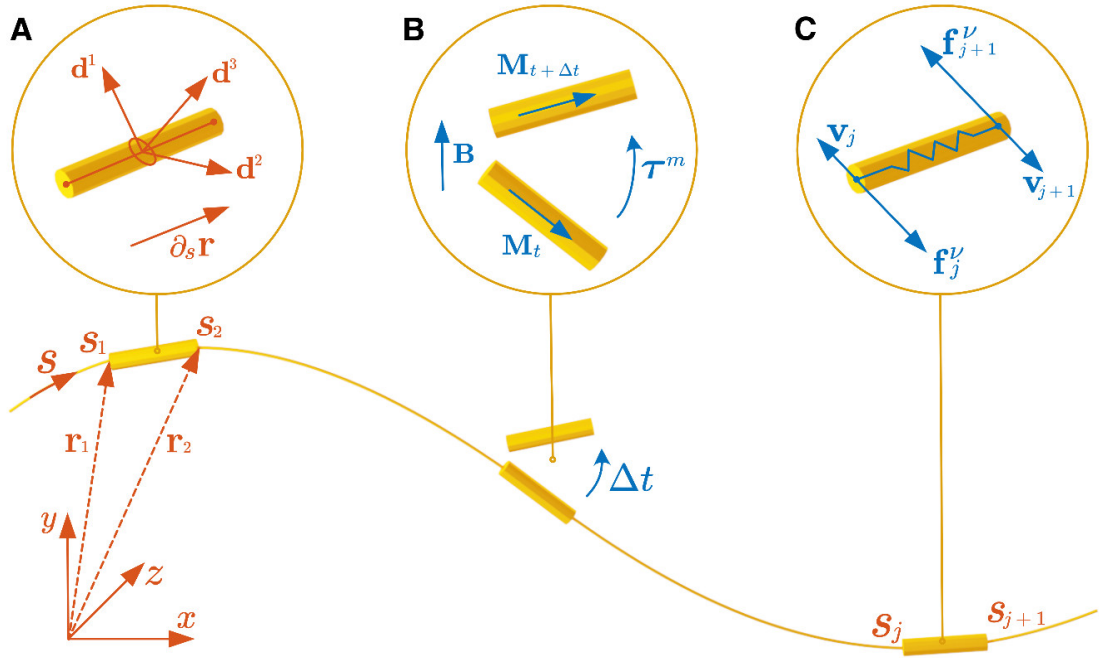

**Fig. 1. Numerical model of magnetic soft robots.** (**A**) Cosserat rod model abstracts a slender structure into a centerline curve and cross-sections. Positional vector **r** is used to describe coordinates of nodes under the Eulerian coordinates. A triad of orthogonal vectors {$\mathbf{d}^1,\mathbf{d}^2,\mathbf{d}^3$} is used to describe the orientation of elemental cross-sections. $\mathbf{d}^3$ is usually chosen to be perpendicular to the corresponding cross-section and is not in the same direction with centerline tangent $\partial_s\mathbf{r}$ due to shear. (**B**) Elements in magnetic soft robots with magnetization **M** are driven by magnetic torques $\boldsymbol{\tau}^m$ to be aligned towards the direction of external magnetic field **B**. **M** is set to be relatively static with the triad {$\mathbf{d}^1,\mathbf{d}^2,\mathbf{d}^3$} so the direction of magnetization will move with the motion of elements. (**C**) Elemental damping forces are modeled to be proportional to the difference of speed between nodes in order to eliminate vibration quickly and at the same time make as little influence on rigid body motion.

To solve the motion problem for an elastic rod with density $\rho$ and cross-section area $A(s)$, we need to utilize Newton's dynamical laws on linear and angular momentum(*42, 47*):

$$\rho(s)A(s)\partial_{tt}\mathbf{r} = \partial_s\mathbf{n}(s,t) + \mathbf{f}(s,t) \tag{1}$$

$$\partial_t\mathbf{h}(s,t) = \partial_s\mathbf{m}(s,t) + \partial_t\mathbf{r}(s,t) \times \mathbf{n}(s,t) + \boldsymbol{\tau}(s,t) \tag{2}$$

In the equations, **n**(*s,t*) and **m**(*s,t*) are the internal force and torque resultants. **h**(*s,t*) is the angular momentum line density. **f**(*s,t*) and **τ**(*s,t*) are external body force and torque line densities.

As for MSRs, they are embedded with fine magnetic particles of some specific magnetization profile. Because of the neglectable volumes and uniform distributions of the particles, when we conduct simulations, we can utilize elemental magnetization to calculate elemental magnetic torques and forces without analyzing the ones undertaken by each particle. When exposed to a uniform magnetic field **B**, an element with magnetization **M** is affected by a magnetic torque $\boldsymbol{\tau}^m$ (*48–49*):

$$\boldsymbol{\tau}^m = V\mathbf{M} \times \mathbf{B} \tag{3}$$

where *V* is the volume of that element. This equation implies that the element tends to be aligned towards the direction of the uniform magnetic field. This magnetic torque can be integrated into the Cosserat model as an external torque in order to simulate MSRs. In this paper, we choose magnetization to be relatively static with the basis $\{\mathbf{d}^1, \mathbf{d}^2, \mathbf{d}^3\}$, as is shown in Fig. 1B.

We adopt a novel dissipation model based on relative speed between adjacent nodes to mimic the high dissipation of soft materials. Intuitively, two adjacent nodes in an MSR can be thought as the ends of a damped spring. When considering damping force, the relative speed between two ends should be taken into account rather than the overall translational velocity of the spring, as is shown in Fig. 1C. So, for two adjacent nodes $n_j$ and $n_{j+1}$ (corresponding arc-length parameters are $s_j$ and $s_{j+1}$) with speeds observed in Eulerian coordinates of $\mathbf{v}_j$ and $\mathbf{v}_{j+1}$, respectively, the damping forces undertaken are

$$\mathbf{f}_j^v = -v(\mathbf{v}_j - \mathbf{v}_{j+1}) \tag{4}$$
$$\mathbf{f}_{j+1}^v = -v(\mathbf{v}_{j+1} - \mathbf{v}_j) \tag{5}$$

where $v$ is a damping coefficient. We found that adding this damping force to our system can significantly eliminate vibration when simulated MSRs perform large-scale deformations or contact with environments and, in the meanwhile, brings only a slight influence on overall rigid body motion. In practice, when models are segmented finely, adjacent nodes $n_j$ and $n_{j+1}$ may have too close velocities. In that case, slightly magnifying the interval between the two nodes helps coarse-tune dissipation. In our work, we choose $n_j$ and $n_{j+4}$ as adjacent nodes to calculate relative speed.

MSRs are also affected by other forces like gravity, contact forces, and friction. For this part, please see (*42*) for modeling details.

We develop our model based on PyElastica (*42, 43*), a numerical package for Python. The original model is for circular-section rods. In order to make the model capable of simulating 2D deformation and movement of band-like MSRs, we make some derivations and equivalent conversions. Please see Materials and Methods: "Conversion from circular-section models to rectangular-section models" section for details.

To validate our model, we conduct simulations of MSRs with sinusoidal magnetization and compare the results with the cases in (*11*) and corresponding FEM simulations. The FEM model is based on Zhao's subroutine (*12, 38*) and gets executed under Abaqus/Standard, widely used in MSR research (*19, 50–52*).

For both FEM and Cosserat rod models, we choose magnetization, Young's modulus, and density of the robots to be 61.3 kA/m, 84.5 kPa, and 1860 kg/m$^3$, respectively, to be in accordance with the values in (*11*). The distribution of magnetic moment density in simulated soft robots is discretized and simplified. The magnetic moment density is assumed to be the same in a particular part. There are no transition zones between two parts with different magnetic moment densities, making little difference in most conditions. The dimensions of robots in simulation models are the same as those in (*11*), which is 3.7 mm×1.5 mm×185 μm.

From the results shown in fig. S1, it is evident that FEM simulations are in better agreement with experiments, which is reasonable because FEM simulations are famous for their accuracies and are usually used for mechanical analysis; however, Cosserat rod simulations are practical enough because they correctly capture the patterns of deformation under different magnetic fields despite the usage of a simple rod structure.

## Intelligent design platform for autonomous learning of MSRs

With the simulation platform described above, information about robot motion can be easily extracted as states and then provided to RL agents for learning. According to Markov property (*53*), a present state must include all helpful information about the past agent-environment interaction that will make a difference for future calculations. However, in practice, too many redundant state components may lead to slower learning processes and even make neural networks diverge. Thus, correctly extracting information from simulated environments to state variables is one of the most challenging yet essential tasks when we implement RL to MSRs.

Because uniform magnetic fields interact with MSRs mainly through torques rather than translational forces, we choose quantities describing rotation as part of state components, such as angles and angular velocities. Besides, we include external magnetic fields to state components as quantities in terms of polar coordinates (angles and amplitudes) to be in better accordance with angular state components of robot motion, as is shown in Fig. 2C and Fig. 2D.

When contacting the ground, MSRs are under gravity, supporting force, and friction. To denote the effect of these forces, we use a simple contact indicator inspired by the ones in OpenAI Gym (*54*). For a node in contact with the ground, the corresponding contact indicator in state components has a base value of 1. Additionally, we add an extra part to describe how strongly the node interacts with the ground, which is proportional to the small deformation of the ground in the simulated environment.

As for positions of MSRs, we only include the height of nodes under the Cartesian coordinates to help agents locate elements and learn moving gaits. We do not include the horizontal distance from the start point because we think animals trying to walk forward do not have to know where they are precisely.

We set our actions to be the increment of external magnetic fields on each time step to make the transition of fields smoother and degrade the effects of inductance. The field increments in two axes range from -0.3 mT to 0.3 mT. The frequency of field updates is 100 Hz in both simulations and experiments.

We set our reward function to be simply proportional to the increase of the horizontal distance of the middle nodes at each time step. Although agents do not know about the exact horizontal positions of the robots, they do know whether the robot is moving forward or not through reward signals.

We use TD3 as our base RL algorithm. TD3 is a modern actor-critic algorithm that has two main parts: an actor for choosing actions according to states and a critic for evaluating state-action pairs. During training, the actor evolves to select actions that are better evaluated by the critic, and the critic evolves to conduct more accurate assessments, as shown in Fig. 2A. Both the actor and the critic use networks to generalize input. For the actor network, states are input, and actions are output; for the critic network, states and actions are input, and Q, the corresponding value estimation, is output. The actor is also known as control policy because it maps states to actions. Network structures of the actor and the critic are shown in Fig. 2E and Fig. 2F. TD3 uses target networks to prevent overestimation and improve performance. Please refer to Materials and Methods: "RL preliminaries" section for detailed explanations.

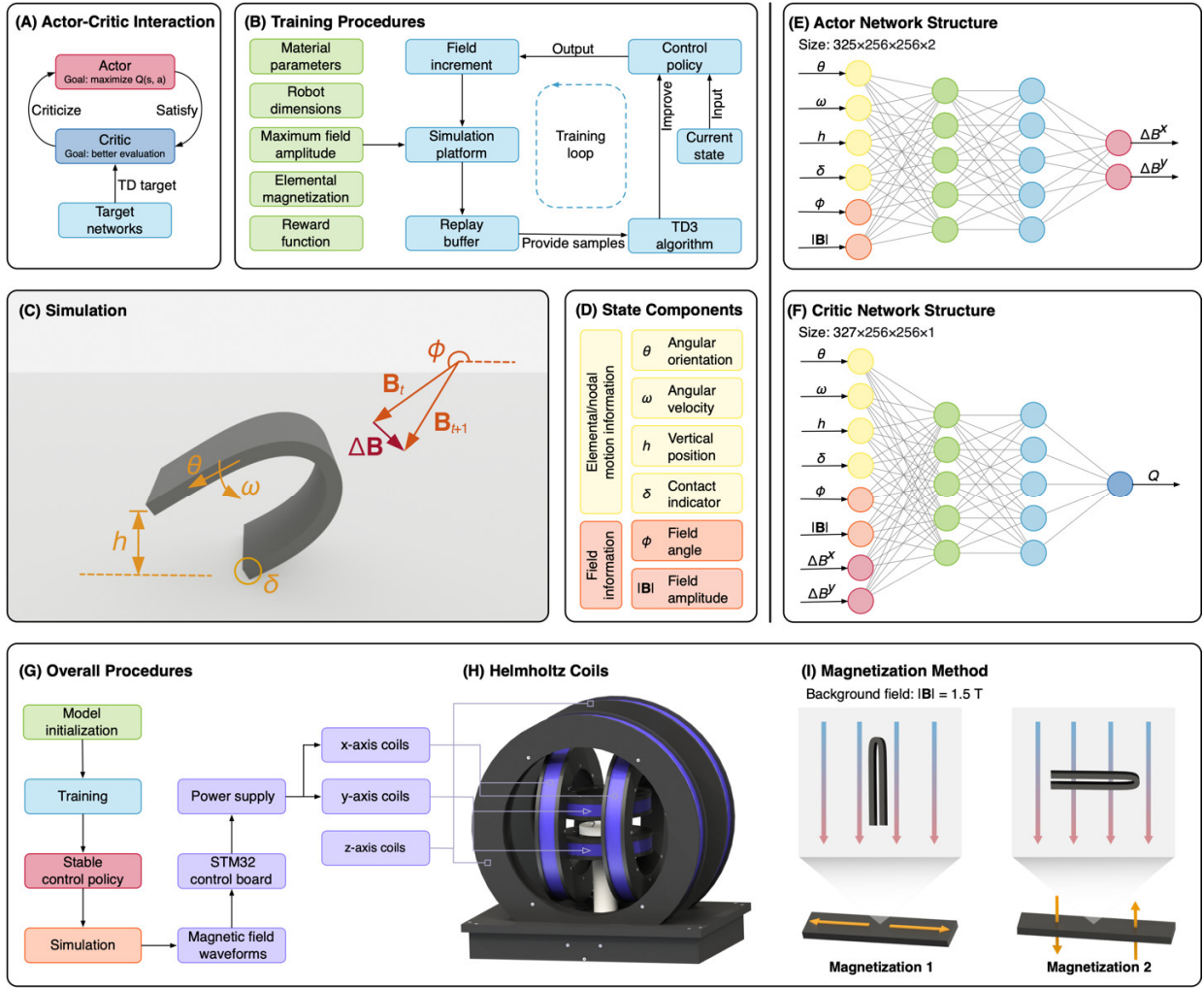

**Fig. 2. Components and procedures of the RL control system.** (**A**) Interaction between different parts of the TD3 algorithm. (**B**) Training procedures for obtaining stable control policies. (**C**) The simulation platform we use. In the picture, we annotate states and actions. In our work, we choose components of field increment ΔB as actions, and quantities involving necessary information of robot motion or magnetic field are defined as states. (**D**) State variables, consisting of angular orientation, angular velocity, vertical position, contact indicator, field angle, and field amplitude. Notice that state variables involving robot motion are defined element-wise or node-wise, so the fineness of segmentation will affect the total size of state variables. (**E**) The structure of the actor network. The actor is used to map states to actions. (**F**) The structure of the critic network. The critic is used to evaluate state-action pairs. (**G**) Overall procedures of our RL control system. After training, we validate the stable control policies in simulations and, at the same time, obtain a set of magnetic fields. The fields are then used to control the power supply via an STM32 control board. (**H**) Helmholtz coils. For now, we use only two pairs of coils. (**I**) Magnetization method. Magnetic soft robots presented in our cases are folded and put into a background field in two directions; therefore, robots with two different magnetization patterns are produced.

As shown in Fig. 2B, to train our RL agents, we first specify physical parameters and some settings about our training tasks and input them to our simulation platform. Then at each timestep, numerical models calculate and update the motion of MSRs. Necessary information produced by the simulation platform is stored in replay buffers, memories of RL agents, as experience transitions. The stored information of all the past interactions with environments is randomly sampled while learning to prevent overfitting. The TD3 algorithm controls learning processes that are also known as the way neural networks update their parameters. A single parameter update does not guarantee the control policy to be better, but when the times of update grow, control policies become increasingly aware of how they should behave. While training, control policies are used to generate actions, in our case, field increments, even if it is still developing. Field increments change magnetic fields of the numerical models, which then take the change into account in the next timestep and continue calculating, updating, storing kinematic information, and so on. Training loops will continue until a predefined maximum timestep is reached.

Deterministic RL algorithms, including TD3, have the problem of run-to-run variance (55). In other words, different agents trained with different random seeds may have very different results, so not all agents will succeed in learning how to control robots to move. In our tests, if we start with 8 random seeds, 3-5 of them will turn out to be stable control policies. The control policies are then used to control simulated soft robots to generate magnetic fields. (Some settings of simulations for generating fields differ from the ones for training. In Materials and Methods: "Miscellaneous training tips", we explain why we set the differences.) The generated field signals are then used to control the power supplies of Helmholtz coils with an STM32 control board, as is presented in Fig. 2G. The Helmholtz coils are used to generate magnetic fields with high uniformity in case of the effects introduced by field gradient, which will add magnetic forces that we do not consider in our numerical models to real MSRs. The field signals decide the corresponding electric currents power supplies provide (See Materials and Methods: "Magnetic actuation system" section). For now, we only use two pairs of our coils to generate magnetic fields in a 2D plane, as can be seen in Fig. 2H. In the future, we plan to upgrade our system to 3D motion control, and at that time, the additional pair of coils will be made into use.

**Learning results under relatively small magnetic fields**
In this and the following sections, we present the learning results of RL agents trained under different conditions. The MSRs we use are of two magnetization patterns, as is shown in Fig. 2I. We also set two different amplitude limitations on magnetic fields. In the cases of this part, maximum amplitudes are set to 4 mT, while in the following part, maximum amplitudes are set to 10 mT.

We fabricate MSRs in a similar manner to our prior work (See Materials and Methods: "Fabrication of MSRs" section), so we adopted the same set of physical parameters in simulation (19). Magnetization, Young's modulus, and density of the simulated soft robots are set to 61.3 kA/m, 84.5 kPa, and 1860 kg/m$^3$, respectively. The dimensions of our robots are 20 mm×8 mm×0.8 mm. The static friction coefficient is set to 0.8, and the kinetic friction coefficient is set to 0.6. As for damping coefficients, we choose them through trials and errors due to the difficulties in theoretical computation and experimental material testing.

When field amplitudes are limited to 4 mT, MSRs with both magnetization patterns learn similar gaits (Fig. 3A, Fig. 3E, movie S1, movie S2). The robots move in a cyclic pattern like crawling, which can be separated roughly into two periods. In the first period, the robots arch up with the distance between two ends shortened. In the second period, the robots stretch with the distance between two ends lengthened, as is shown in Fig. 3B and Fig. 3F. In both cases, two ends of the robots keep in touch with the ground, and the strength of touching varies as they move. About the time when robots reach their maximum height, the contact strength of the back ends becomes obviously larger than the front ends. From the position and velocity plots (fig. S4), the moving gaits of both robots are similar to "anchor push-anchor pull" locomotion strategy of inchworms (56).

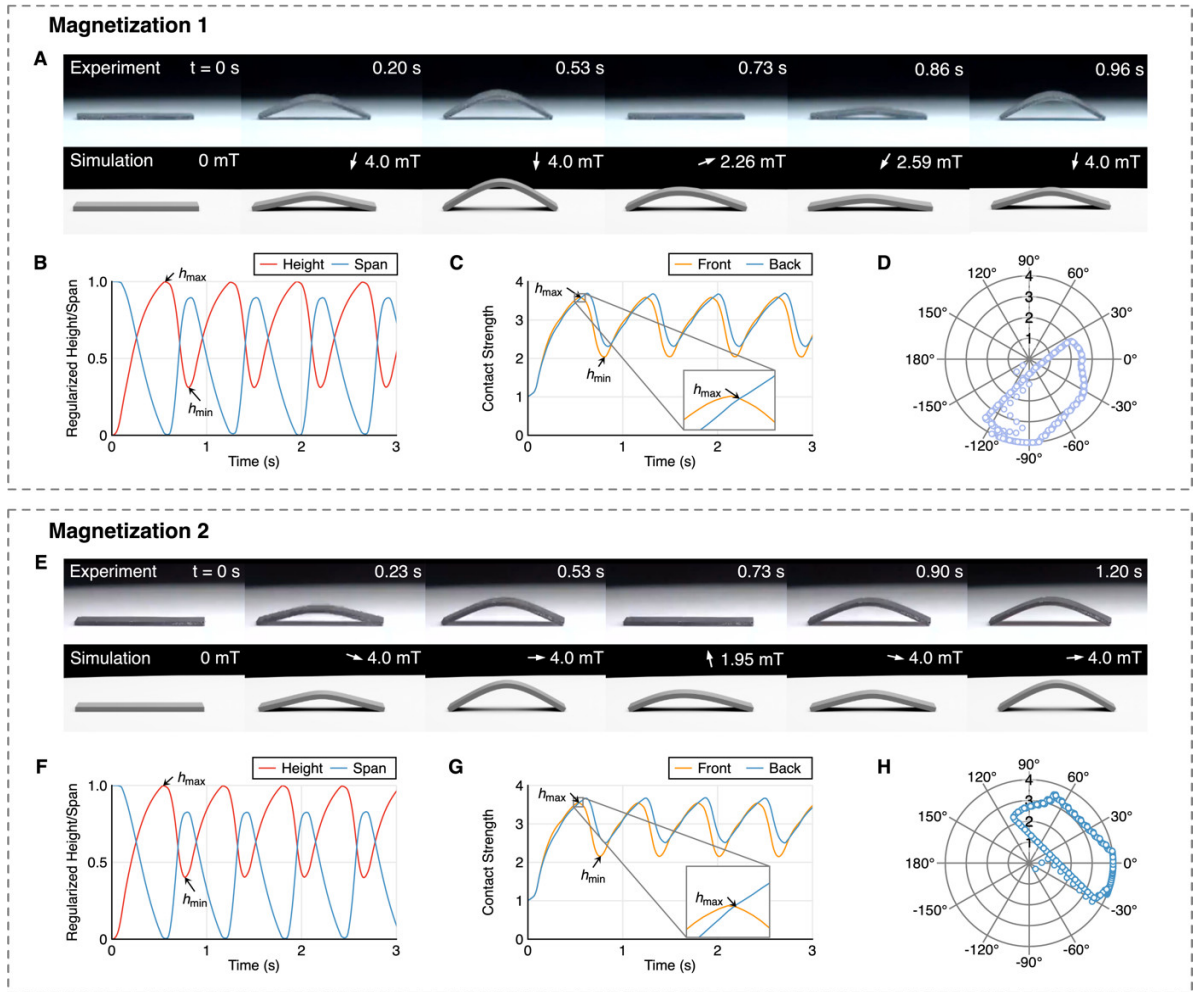

**Fig. 3. Learning results under relatively small field amplitude.** Notice that the plots are from data in simulations. The dimensions of our robots are 20 mm×8 mm×0.8 mm. (**A**) Comparisons between simulation and experiment of the robot with magnetization 1. Directions of magnetic fields are indicated with a white arrow, and amplitudes of fields are annotated in number. The real soft robot reaches its maximum height about 0.3 s faster than the simulated one because our damping model leads to some overdamping effect on rotation, but overall moving gaits between the two are in good agreement. See movie S1 for more. (**B**) Regularized height or span of the robot in 0 to 3 s. When the span reaches its minimum, the height reaches its maximum, and vice versa. (**C**) Strength of the front end and the back end contacting the ground in 0 to 3 s. When the height of the robot reaches a maximum, the contact strength of the back end becomes larger than the front end. (**D**) Scattered field plots of applied to the robot in 0 to 10 s. (**E**) Comparisons between simulation and experiment of the robot with magnetization 2. See movie S2 for more. (**F-H**) The plots of the robot with magnetization 2 corresponding to Fig. 3B-D. The learned moving gait is very similar to the one of magnetization 1. Two scattered plots of fields share similar contour shapes but have different orientations, which indicates similar torques among the two robots.

As is shown in movie S1 and movie S2, experiments and simulations are consistent, except that the real soft robots respond to fields more rapidly. This is because our damping model is not perfect, as can be explained using an example of a rigid rod. If the rigid rod rotates around its middle point, it will keep rotating in the real world if no other forces are given, but in our numerical model, the rod cannot rotate uniformly due to the damping we add. However, as is observed in experiments, in the processes of crawling, real robots reach their maximum height about 0.3 seconds faster than simulated robots, which is acceptable in our cases.

Since the magnetization patterns of the two robots are not the same, the learned control policies generate different magnetic fields in the two cases. We plot discrete field points in polar coordinates as is shown in Fig. 3D and Fig. 3H. It is evident that contour shapes in the two plots are very much alike, and if we rotate the plots in Fig. 3D anticlockwise about 90 degrees, we can get a similar plot as Fig. 3H, which makes absolute sense because the orientation difference of fields is the same as the orientation difference of magnetization. By rotating fields, two RL agents generate similar magnetic torques on both robots and thus, control both robots to move forward in similar gaits. This is interesting because RL agents do not even know about the magnetization patterns, but they learn from interaction, and produce results meaningful in mechanics.

**Learning results under relatively large magnetic fields**

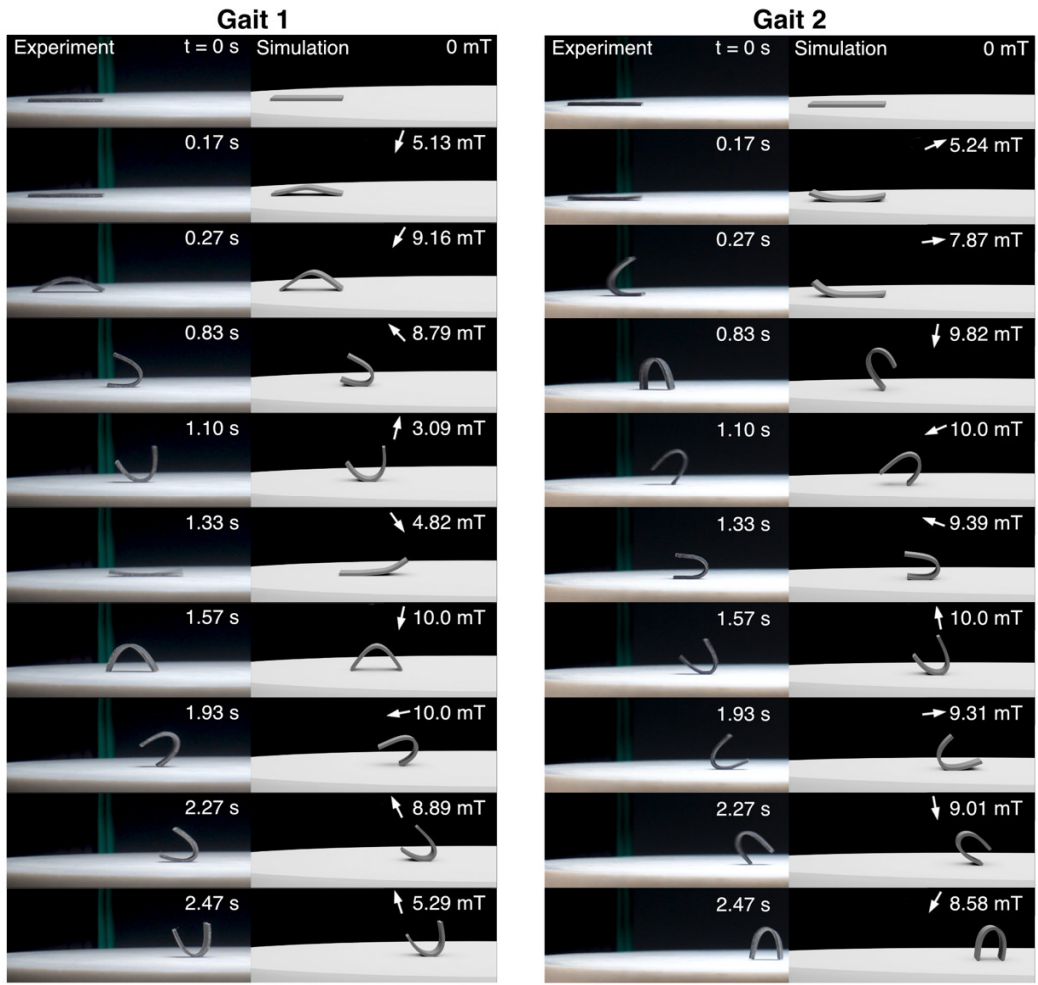

**Fig. 4. Comparisons between simulations and experiments of the robot with magnetization 1 under relatively large field amplitude.** Two different gaits are learned due to run-to-run variances of TD3. The experiments show good agreements with simulations with the exception of highly dynamic conditions at the beginning of gait 2 due to overdamping on rotations. See movie S3 and movie S4 for more.

In this part, we present cases with a maximum field amplitude of 10 mT. Because when magnetic fields have larger amplitude, soft robots have much larger deflection and more various deformation patterns than under relatively small fields, the learning results show diverse moving gaits. For robots with magnetization 1, two very different gaits are learned, as is shown in Fig. 4. The first gait can be separated into about 3 periods (fig. S5E). In the first period, the robot arches up and, at the same time, keeps the orientation straight down. In the second period, the robot rotates clockwise, a little less than 180 degrees. In the third period, the robot unfolds, then again, followed by arching. The second gait is simply clockwise rolling (fig. S5F).

The robot with gait 2 moves quicker than the robot with gait 1. Considering that the defined reward function is simply proportional to the distance the middle points move forward, the second gait is better than the first. However, the velocity of the midpoint in the second gait is much more fluctuant than the velocity in the first gait (See Materials and Methods: "Additional data of the cases" and fig.S5G). RL agents may fit into different local optimum peaks of value estimations, which results in different control policies and moving gaits.

As for the robot with magnetization pattern 2, we find that when deflection is large, the two ends will easily get attracted to each other in experiments. This is because the magnetization polars of the two ends are opposite, leading to magnetic attraction between the ends, which has not been considered yet. We show this failure case in fig. S2 and fig. S6 for completeness.

**Analysis of RL agents**
Different agents choose different actions under similar states because their evaluations of actions differ. Here we plot the Q-value distribution for different actions in Fig. 5.

When small magnetic fields are applied, we choose the timesteps at which robots with both magnetization patterns reach their maximum or minimum heights. When the robot with magnetization pattern 1 reaches its highest point, the best field increment estimated by the critic is that for both axes, field components get incremented by 0.3 mT, which corresponds to the actual field increment shown in field waveforms (See fig. S4E). Similarly, in most cases, actual chosen actions are the ones with the highest Q values estimated by critics, except when the robot of magnetization 2 reaches its lowest point, as is shown in Fig. 5B. The actor chooses the field increment to be 0.3 mT in $B_x$ and -0.3 mT in $B_y$, while the action with the best Q-value is about -0.15 mT in $B_x$ and -0.3 mT in $B_y$, which indicates that the actor does not fully satisfy the critic under the particular state. In fact, actor-critic algorithms do not guarantee that actions chosen by the actor always have maximum Q values in every step, which is quite different from algorithms for discrete action spaces like DQN (*34*). Approximating control policies using neural networks will lose some accuracy in choosing the best actions; however, thoroughly finding the best actions is unrealistic due to the vast amount of computation for problems with continuous action spaces. And on the other hand, sometimes choosing suboptimal actions still help solve tasks, like the cases presented in our work.

When large magnetic fields are applied, we pick the starting condition shared by both gaits because moving gaits are so different. Because the agent of gait 1 tries to arch the robot up while the agent of gait 2 wants to make the robot roll, the Q-value estimations are not alike, even though they are trained under same restrictions.

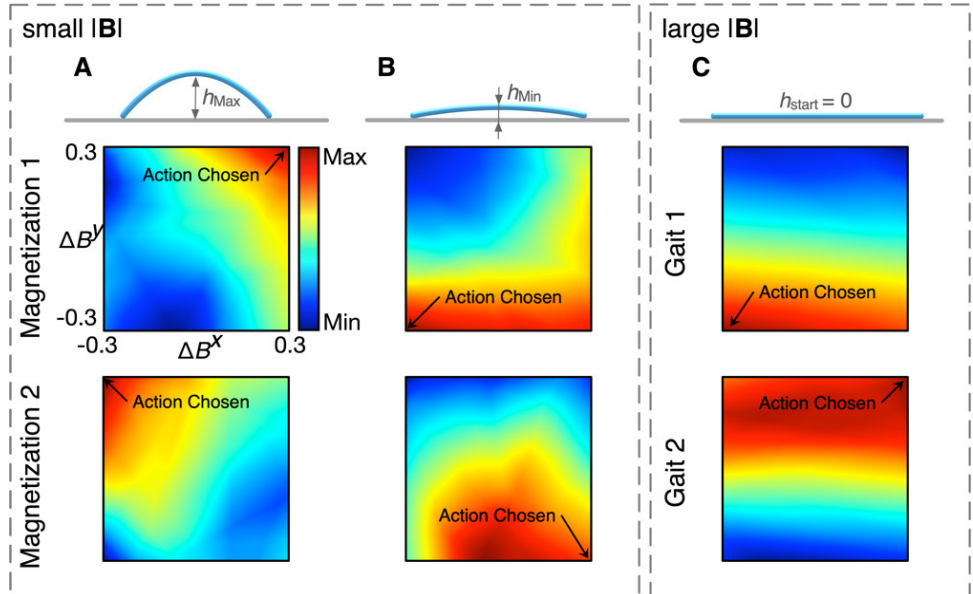

**Fig. 5. Q-value distributions for different actions estimated by different agents.** We get these heat maps by inputting uniformly distributed actions to critics under certain states. The axes are in the unit of mT. In the plots, the points annotated by "Action Chosen" are the actual field increment chosen by actors. (**A**) Q-value distributions when robots reach their highest point. Two agents trained under different magnetization patterns both choose different actions that correspond to points with a maximum Q value. (**B**) Q-value distributions when robots reach their lowest point. The action selected by the agent of magnetization 2 does not have the highest Q-value. (**C**) Q-value distributions of agents corresponding to two gaits shown in Fig. 5. We choose the specific state to be at the starting position because it is shared by both gaits.

## Discussion

In this work, we present one of the first intelligent design systems for MSRs and demonstrate the potential of RL to achieve autonomous manipulation of these robots. Specifically, we develop a new simulation method based on Cosserat rod models for predicting dynamic behavior of MSRs. The simulations are in good agreement with experimental results for both static and dynamic cases, which provides us with an approach to open-loop control of real MSRs with the magnetic fields generated in simulations. Based on that, we introduce deep RL algorithms to solve the problem of magnetic actuation for soft robots inversely without human guidance. We extract key kinematical features of simulated MSRs as state variables input to RL agents that then learn from interaction with environments and successfully generate valid control policies and fields to control MSRs to move. We present several cases involving different magnetization and different field limitations, demonstrating the generality of our method. We believe that our work can help accelerate research in magnetic soft robotics by introducing the brand-new inverse-design method. Owing to the difficulties in analyzing the highly nonlinear motion of soft robots, previous researchers mainly use intuition to design and control. Now, with the help of modern RL algorithms, researchers can focus on what they want to fulfill rather than how to do so and let RL agents automatically design proper parameters.

Our work can be further improved from the following aspects. First, the single-rod model still owns limitations on generality. For further improvement, we plan to combine multiple rods together to formulate more complex shapes and deformation patterns. Besides, the dissipation model still needs improving due to the overdamping it brings. Additionally, we think it necessary to combine more advanced deep learning structures with our system and extend the scenarios to more general tasks. Lastly, we aim to use sensors to introduce feedback by combining sensors (*57*) or building an image capturing system (*58*) in our experiments and formulate a closed-loop control system.

## Materials and Methods
### Robot fabrication
The magnetic-responsive material used in the experiment is a mixture of Ecoflex 00-10 polymer matrix (Smooth-On Inc.; density: 1.04 g/cm$^3$) and neodymium (NdFeB) magnetic powders (MQP-15-7, Magnequench, density: 7.61 g/cm$^3$) with a weight ratio of 1:1. To ensure uniform particle dispersion, the blend was first mixed at 2000 r.p.m for 1 min by the planetary mixer (AR-100, THINKY Corp), followed by defoaming at 2,200 r.p.m. for 1 min. Then the thoroughly blending magnetic slurry was scraped into the prepared mold. After thermal polymer curing (Room temperature, 4 hours), the demolding sample was magnetized by a pulsed magnetic field (about 1.5 T) generated by the pulsed magnet (*59*) to program the internal magnetization.

### Magnetic actuation system
RL agents control field increments at a frequency of 100 Hz, so we extract 100 discrete points of magnetic fields per second from simulations and form a CSV file for each task. The file contains *x*-, *y*-, and *z*-axis fields (in our cases, *z*-axis fields are always 0). An STM32F407 microcontroller is used to read the file, conduct related calculations, and then output continuous voltage signals through a DAC8563 module. The voltage signals are amplified by three power amplifiers (HEA-500G, HEA-200C, HEAS-50 for *x*-, *y*- and *z*-axis fields, respectively), generating currents in 3D Helmholtz coils (3HLY10-100). The field amplitudes of three axes are proportional to corresponding currents in coils. To maintain reliable and repeatable experiments, we regularly calibrate the magnetic actuation matrix inside the workspace, i.e., the mapping between the applied electric current and the generated magnetic field.

### RL preliminaries
This section aims to provide a relatively intuitive RL introduction for interested readers from different areas and explain the reasons for choosing TD3 as our base RL algorithm.

(1) Core concepts of RL

According to Markov Decision Processes (MDPs), the learning problem can be framed into several parts. The learner, as well as the decision-maker, is called the *agent*. The things the agent interacts with make up the *environment*. At each time step *t*, the agent selects an *action a* based on the environment's current *state s*. The term *policy* π is used to describe the mapping between *a* and *s*. After that time step, the environment sends a *reward r* to the agent, which is usually correlated with a goal the agent is meant to achieve and turns into a new state *s'*.

For optimal control problems, we aim to find the best policy $\pi_*$ that can most likely get the best results. In order to achieve that, the optimal action-value function $Q^*(s, a)$ that evaluates actions based on their expected *return* needs to be calculated. Return means the weighted sum of all future rewards, and in definition, $Q^*(s, a)$ is expressed as:

$$Q^*(s,a) = \max_{\pi} \mathbb{E}[r_t + \gamma r_{t+1} + \gamma^2 r_{t+2} + ... | s_t = s, a_t = a, \pi] \quad (6)$$

However, the definition formula can be hard to compute. The solution to this problem is the framework of generalized policy iteration and temporal difference update. Simply speaking, generalized policy iteration means that as long as we take steps between a better action-value evaluation and a better policy improvement, we can eventually get the true optimal action-value function $Q^*(s, a)$ no matter what the start values are. On the other hand, temporal difference updates enable us to use the action-value functions of the next state $s'$ as a reference to update the action-value functions of the current state $s$. If the action-value function starts arbitrarily as $Q(s, a)$, the update rule can be expressed as:

$$Q(s,a) \leftarrow Q(s,a) + \alpha(r + \gamma \max_{a'} Q(s',a') - Q(s,a)) \quad (7)$$

This update rule is also known as Q-learning. In the equation, $\max_{a'} Q(s',a')$ is the best possible action-value function of all the next state-action pairs, and it has to be multiplied by discount factor $\gamma$ to ensure convergence. The iteration can be understood as a step towards the optimal action-value function $Q^*(s, a)$, with a step size of $\alpha$. $r + \gamma \max_{a'} Q(s',a')$ is called TD target, which points out the direction and target of iteration steps. The iterating $Q$ value will eventually converge to real $Q^*(s, a)$ (60).

Temporal difference update introduces a method for updating action-value functions between time steps. This is extremely meaningful to practical RL algorithms because lots of real-life tasks are non-episodic, without a natural terminal state to mark the end position that is indispensable for calculating a return value.

(2) Development of practical deep RL algorithms

The original RL framework fits naturally into problems with discrete states and actions. However, for real-life issues, states are seldomly discrete; instead, state variables are usually continuous and get restricted by some ranges. Thus, traditional methods that use tables to store action-function values are impractical to solve problems in industries.

Deep Q-learning uses a neural network as a function approximator. In other words, the neural network automatically extracts useful features from the input data for solving the control problems, and how to extract useful features (i.e., the parameters of neural networks) is learned by interacting with the environment. Thus, continuous values of input states can be generalized by the nonlinear function presented by the neural network.

In the previous discussion of Q-learning, TD target is the target of iterating action-value function, which also gives out an implication of the loss function for the neural network of deep Q-learning:

$$L(\theta) = \mathbb{E}[(r + \gamma \max_{a'} Q(s',a';\theta) - Q(s,a;\theta))^2] \quad (8)$$

$r + \gamma \max_{a'} Q(s',a';\theta)$ is the TD target of deep Q-learning. $\theta$ represents the parameters of the action-value Q network. The neural network uses gradient descent (61) to update its parameters and hopes to eventually make the loss function a local minimum. In other words, in the ideal cases, the action-value function represented by the neural network will converge to be equal to the TD target. However, this naive version of deep Q-learning is unstable and can hardly to applied directly.

One of the most important advancements in modern RL is DQN (34). The DQN paper presents the first artificial agent capable of learning to excel at a diverse array of challenging tasks. For example, the agent achieved human-level control on the Atari games, which depended on an end-to-end structure that used game images as input.

Two critical innovations of DQN greatly enhance the stability of using neural networks to approximate action-value functions: replay buffer and separate target network. The replay buffer is used to store separated experience transitions $e_t = (s_t, a_t, r_t, s_{t+1})$ observed when the agent is interacting with the environment. Then, when learning, the agent samples experience transitions $e_t$ from the replay buffer $D_t = \{e_1, e_2, e_3, \ldots, e_t\}$ randomly and uses them to conduct gradient descent. Thus, the observation sequence can be broken apart, and the agent is less prone to overfit the data in a specific time span. The second innovation, the delayed-update target $Q'$ network, is designed to make the temporal difference target in the update rule more consistent. Therefore, the neural networks in RL agents are less prone to diverge.

The loss function for $Q$ network is

$$L(\theta) = \mathbb{E}_{(s,a,r,s') \sim U(D)}[(r + \gamma \max_{a'} Q'(s',a';\theta') - Q(s,a;\theta))^2] \tag{9}$$

in the equation, $r + \gamma \max_{a'} Q'(s',a';\theta')$ means that the TD target is no longer calculated based on the $Q$ network, as is in deep Q-learning. Instead, it uses the delayed-update target $Q'$ network, whose parameters are represented as $\theta'$. $\theta'$ is not updated by gradient descent or other learning methods. It simply copies the values of $\theta$ every $C$ steps where $C$ could be a hyperparameter to tune. The expectation is calculated and evaluated based on the experience transitions from the replay buffer, as is shown by $(s,a,r,s') \sim U(D)$. For simplicity, the following discussions omit $(s,a,r,s') \sim U(D)$ in equations. As long as an algorithm utilizes a display buffer, it will learn from samples of experience transitions.

(3) Modern actor-critic algorithms: DDPG and TD3

DQN can solve problems with continuous state space and discrete action space; however, in robotics, the agent we aim to control usually has a continuous span of action space. In order to solve this problem, the DDPG algorithm (*62*) brought innovations from DQN to the actor-critic framework.

Actor-critic algorithms differ according to different purposes or under different prerequisites. For example, the policy we want to approximate may be stochastic or deterministic, which also changes other parts of the algorithms. Here we focus on the basic structure that DDPG built upon, which utilizes a network to approximate the deterministic policy $\pi(s)$, besides a network to approximate the action-value function $Q(s, a)$. The former network is called the *actor* network, and the latter is called the *critic* network. Two networks aim to achieve different goals, but together they make the algorithm converge to an optimal policy.

Because policy $\pi$ describes the mapping between actions and states, the actor network takes state values as input and calculates action values as output. The goal for the actor network is to choose the best policy, as is implied by its loss function:

$$L(\theta^\pi) = -\mathbb{E}[Q(s, \pi(s;\theta^\pi); \theta^Q)] \tag{10}$$

In the equation, the minus sign is added in the beginning because the actor aims to maximize $\mathbb{E}[Q(s, \pi(s;\theta^\pi);\theta^Q)]$ part, and in most modern neural network frameworks, loss function and gradient descent are the default settings rather than the reward function and gradient ascent. $\theta^\pi$ refers to the parameters in the actor network, and $\theta^Q$ refers to the parameters in the critic network. $\pi(s;\theta^\pi)$ means that the action is an output from the actor network that represents the policy $\pi$. The meaning of this equation can be quite straightforward: we update the parameters of the actor neural network so that the actor (or policy) can choose better actions that achieve a higher action-value estimated by the critic.

On the other hand, the loss function for the critic network is

$$L(\theta^Q) = \mathbb{E}[(r + \gamma Q'(s', \pi(s';\theta^\pi);\theta^Q) - Q(s,a;\theta^Q))^2] \tag{11}$$

which can also be viewed as two main parts. $r + \gamma Q(s', \pi(s'; \theta^\pi); \theta^Q)$ is the TD target, and $Q(s, a; \theta^Q)$ is the action value estimated by the critic. The parameters of the critic network get adjusted so that the estimated action value gets closer to the TD target, just as in deep Q-learning.

Besides, DDPG adopts target networks and replay buffer to enhance the performance of the actor-critic structure. DDPG uses a target $Q'$ network along with a target $\pi'$ network to calculate the TD target. Thus, (11) should be changed to

$$L(\theta^Q) = \mathbb{E}[(r + \gamma Q'(s', \pi'(s'; \theta^{\pi'}); \theta^{Q'}) - Q(s, a; \theta^Q))^2] \tag{12}$$

while the loss function for the actor network remains the same. In the equation, $\pi'(s'; \theta^{\pi'})$ means that the policy for calculating target action value $Q'$ is derived from the target $\pi'$ network and $Q'(s', \pi'(s'; \theta^{\pi'}); \theta^{Q'})$ means that the target action value $Q'$ in the TD target is derived from target $Q'$ network.

The update rule for the target networks is different from the one in DQN. The weights in the target $\pi'$ network and target $Q'$ network are updated by slowly tracking the weights in the $\pi$ network and $Q$ network using exponentially weighted average (*63*). The weights should be updated slowly to prevent divergence.

TD3 further improved stability and performance by introducing three main improvements to DDPG in order to partly eliminate variance and bias brought by function approximation error.

The first improvement is target policy smoothing regularization, which aims to reduce the variance of the TD target. Target actions are smoothed (*37*) by adding a Gaussian noise so that similar actions would have similar values and the trained deterministic policies are less prone to overfit to narrow peaks of value estimation. Thus, the TD target of TD3 can be expressed as:

$$y = r + \gamma Q'(s', \pi'(s'; \theta^{\pi'}) + \epsilon; \theta^{Q'}) \tag{13}$$

where $\epsilon$ is a noise sampled from a clipped Gaussian distribution.

The second improvement is a pair of critics to address overestimation bias (*37*), which is built on Double Q-learning (*64*). In TD3, two separate critic networks $Q_1$ and $Q_2$, along with two target critic networks $Q'_1$ and $Q'_2$, are used. When updating parameters, two target values are calculated

$$y_1 = r + \gamma Q'_1(s', \pi'(s'; \theta^{\pi'}) + \epsilon; \theta^{Q'_1}) \tag{14}$$
$$y_2 = r + \gamma Q'_2(s', \pi'(s'; \theta^{\pi'}) + \epsilon; \theta^{Q'_2}) \tag{15}$$

The TD target used in the loss functions for both critic networks is

$$y = \min(y_1, y_2) \tag{16}$$

and the loss functions for $Q_1$ and $Q_2$ are

$$L(\theta^{Q_1}) = \mathbb{E}[(y - Q_1(s, a; \theta^{Q_1}))^2] \tag{17}$$
$$L(\theta^{Q_2}) = \mathbb{E}[(y - Q_2(s, a; \theta^{Q_2}))^2] \tag{18}$$

The third improvement is delayed updates on the actor network π along with target networks π′, $Q'_1$ and $Q'_2$. Similar to DDPG, the loss function for π is

$$L(\theta^\pi) = -\mathbb{E}[Q_1(s, \pi(s; \theta^\pi); \theta^{Q_1})] \tag{19}$$

Here we choose $Q_1$ as the value estimator in accordance with the original TD3 paper. As can be seen, the actor network aims to maximize the estimated action value. However, if the estimated value itself is inaccurate with high variance, the calculated policy will be more inclined to diverge. Using delayed policy updates means that we update our actor network and target networks at a lower frequency than the critic networks so that we can stabilize the estimated action value before conducting a policy update. In this way, the performance of the actor improves though with fewer updates.

**Conversion from circular-section models to rectangular-section models**
Because we develop our simulation platform based on PyElastica (*42, 43*), an open-source Cosserat rod package for Python, and the original model is for rods with circular sections, we have to make some conversions in order to make it valid for simulations of bar-like MSRs.

The main forces that determine deformations of MSRs consist of magnetic torques, gravity, and elastic forces. Magnetic torques and gravity activate deformations, so we call them "active forces/torques" in this section. In contrast, the elastic forces will be generated only when deformation happens, so we call them "passive forces/torques".

The original rod model in PyElastica supports fully 3D deformations, including 6 deformation modes; however, in our work, we care only about 2D deformation and movement of MSRs so that the deformation modes can be simplified to 3 deformation modes: bending about tangents, shearing along tangents, and stretching along normals. The modes and corresponding elastic strains, rigidities, and loads are shown in table S1.

Imagine a circular-sectional rod with a radius of *r* and a rectangular-sectional rod with a width of *w* as well as a height of *h*, as shown in fig. S1. Take small segments with a length of $\Delta l$ from both rods. Magnetization of both parts is **M**, and density is *ρ*. When applied with an external magnetic field **B**, magnetic torques undertaken by the two segments are

$$\boldsymbol{\tau}^{\text{cir}} = V^{\text{cir}} \mathbf{M} \times \mathbf{B} \tag{20}$$
$$\boldsymbol{\tau}^{\text{rec}} = V^{\text{rec}} \mathbf{M} \times \mathbf{B} \tag{21}$$

where *V* stands for volume of the segments, and superscripts indicate shapes of cross-sections. The volumes are calculated as

$$V^{\text{rec}} = wh\Delta l \tag{22}$$
$$V^{\text{cir}} = \pi r^2 \Delta l \tag{23}$$

So, for the two segments, the torques undertaken are proportional to the areas of sections.

Besides, the gravities undertaken by both of the segments are also proportional to the areas of sections, for

$$\mathbf{G}^{\text{rec}} = wh\Delta l \rho \mathbf{g} \tag{24}$$
$$\mathbf{G}^{\text{cir}} = \pi r^2 \Delta l \rho \mathbf{g} \tag{25}$$

During the conversion, we aim to keep strains equivalent in both circular and rectangular rods. Say combined active forces and torques are **F** and **τ,** respectively, the strains generated are then

$$\Delta \boldsymbol{\sigma} = \mathbf{F} / S \qquad (26)$$
$$\Delta \boldsymbol{\kappa} = \boldsymbol{\tau} / B \qquad (27)$$

where $S$ and $B$ are rigidities. So, when **F** and **τ** are proportional to the areas of sections, if rigidities are proportional to the areas of cross-sections as well, the strains will be the same.

Stretching rigidity $S_n=EA$ and $E$ is the same for both segments, so it satisfies the requirement. As for bending rigidity $B_t=EI_t$, we have to solve the equation

$$\frac{wh^3}{12} / \frac{\pi r^4}{4} = \frac{wh}{\pi r^2} \qquad (28)$$

where $wh^3/12$ is the second moment of inertia for the rectangular segment and $\pi r^4/4$ is the second moment of inertia for the circular segment. $wh/\pi r^2$ is the ratio of two areas of cross-sections. From the equation, we can get $r=\frac{\sqrt{3}}{3}h$. Therefore, when the radius of the circular segment is equal to $\sqrt{3}/3$ times the height of the rectangular segment, bending rigidities of both segments are proportional to the areas of sections.

As for shear rigidity $S_t=\alpha_c GA$, if we set the constant $\alpha_c$ to 1 rather than 4/3 of the original model, we can meet the requirement.

To conclude, we can use a rod model with a circular cross-section to simulate the 2D motion of a bar-like MSR if we set the radius equal to $\sqrt{3}/3$ times the height of the robot and set the constant $\alpha_c=1$.

**Miscellaneous training tips**
When training, we set a maximum time in order to reset simulations regularly in case agents get stuck in some situations. In our work, we set the maximum time to be 20 s.

But notice that the moving of MSRs should not be modeled as an episodic task. According to definitions, terminal states have a state value of 0, so if the problem is modeled as an episodic task and the terminal states correspond to the states when time hits 20 s, the RL agents may get confused while learning. This is because random states that include information only about motion and fields may be marked as terminal states, and the terminal states may also appear as non-terminal states before time hits 20 s, which leads to conflicting directions when the critic updating its network parameters. We do not include time as a state component because we think specific time is not essential in the period of moving forward.

In order to model the problem as a continuous task and at the same time utilize the benefits of a maximum timestep, we skip the experience transitions corresponding to the maximum timestep and choose not to add them to replay buffers.

Besides, in order to diversify samples stored in replay buffers and prevent RL agents from overfitting. At the beginning of each episode, we set random initial magnetic fields within the range of maximum amplitude. But notice that due to inductance, the current in real Helmholtz coils must start from 0, so when running simulations for generating fields after stable control policies are obtained, we cancel the settings of random initial fields and make the fields start from 0 in order to generate fields that can be directly loaded.

Lastly, because RL is highly commanding in computational resources, we figure out a way to shorten training periods. We notice that the PyElastica numerical package has difficulties dealing with small-scale soft robots. The smaller the soft robots are, the smaller calculation timesteps have to be, which will significantly elongate total calculation time. We think it may be due to the small moment of inertia of elements in small-scale soft robots, which leads to enormous velocity increments in a single timestep. In order words, if timesteps are large, velocity growth in a single timestep may easily exceed the limits of NumPy. We find that if we increase the density of the robots while at the same time inverse-proportionally decrease gravitational acceleration, the gravity as well as deformation of robots will remain the same, and timesteps can be set to larger values. In our case, when we set density to be 10.5 times larger and gravity acceleration to be 10.5 times smaller, calculation timesteps can be set from $8\times10^{-6}$ s to $1\times10^{-4}$ s, which brings about 10 times faster calculation.

However, this method affects the dynamic responses of our numerical model. In order to solve this problem, after we train our RL agents for about $1\times10^5$ training steps in the inaccurate simulation environment, we will refine the stable control policies in a more accurate simulation environment with regular density and gravity for about 1000 steps. And when we generate magnetic fields, the policies are run in accurate simulation environments. We use a 16-core Intel i7 for our training. A single period consisting of training, refining, and generating waveforms takes less than 24 hours, while 6 independent seeds can be trained simultaneously.

For more specific parameters or hyperparameters, please see our code.

**Additional data on the cases**
(1) Learning results under small field amplitude
We run 8 random seeds for each task, and 3 to 5 of them will successfully generate stable control policies. For RL agents trained under small field amplitude, stable control policies show similar gaits, so we pick the ones with the highest average scores as the cases to present. Average learning curve across the 8 seeds and the learning curve of the case we present of the robots with two magnetization patterns are shown in fig. S2A-B, F-G. As can be seen, the average learning curve of magnetization 2 shows less reward than the one of magnetization 1, which may be due to randomness. The learning curves of the cases we present in our work show similar average rewards for both robots. We extract the lateral positions of the front, middle and back nodes and plot them in fig. S2C, H. As can be seen, middle nodes move much stabler and the others, which is probably due to the fact that we set lateral positions of the middle nodes as reference for reward functions. We use the plots of positions to generate the velocity plots by differentiation, as shown in fig. S2D, I. Obviously, the velocities of middle nodes are much less fluctuant than other nodes and have the least negative velocities. When robots arch up to the highest point, the back nodes move forward while the front nodes negligibly move backward; when robots unfold to the lowest point, the front nodes move forward while the end nodes negligibly move backward. The middle nodes almost always have positive velocities. The peak velocities of the front nodes are much larger than the others. The magnetic field waveforms are shown in fig. S2E, J. We can see that the field amplitudes of the two cases are very similar, but $B_x$ and $B_y$ are in very different shapes.

(2) Learning results under large field amplitude
For RL agents trained under large field amplitude, stable control policies show different gaits, so we pick two typical gaits as the cases to present in our work. The corresponding learning curves are shown in fig. S3A-C. The second gait has larger average rewards than the first gait. We pick the opening orientation angle $\beta$ as a quantity to depict the phases of motion. $\beta$ is defined as the angle between the *x*- axis and the vector pointing from the middle node to the middle of two ends, as is shown in fig. S3D. The change of $\beta$ in two gaits is plotted in fig. S3E and fig. S3F. It can be seen that in gait 1, the robot first arches up, then rolls for less than 180°, and then unfold. The motion can be separated into 3 phases in total. But in gait 2, the soft robot rolls all the time. In fig. S3G, the lateral velocities of the middle nodes in two gaits are plotted. It can be seen that the velocity in gait 2 is much bumpier than the one in gait 1. We plot the trajectories of the middle nodes in the $x$-$y$ plane in fig. S3H and fig. S3I. In fig. S3J-L, we plot the magnetic waveforms. It can be seen that the field in gait 2 is more like a rotating field and has a larger amplitude than the one in gait 1.

For the MSR with magnetization pattern 2, when applied with large field amplitude, two ends of the robot will stick to each other in experiments. So, we only show the learning curves and field waveforms in fig. S4 and do not conduct any further analysis for this failure case.

**References**
1. M. Li, A. Pal, A. Aghakhani, A. Pena-Francesch, M. Sitti, Soft actuators for real-world applications. *Nat. Rev. Mater.* **7**, 235–249 (2022).
2. M. Ilami, H. Bagheri, R. Ahmed, E. O. Skowronek, H. Marvi, Materials, actuators, and sensors for soft bioinspired robots. *Adv. Mater.* **33**, 2003139 (2021).
3. N. S. Usevitch, Z. M. Hammond, M. Schwager, A. M. Okamura, E. W. Hawkes, S. Follmer, An untethered isoperimetric soft robot. *Sci. Robot.* **5**, eaaz0492 (2020).
4. D. Drotman, S. Jadhav, D. Sharp, C. Chan, M. T. Tolley, Electronics-free pneumatic circuits for controlling soft-legged robots. *Sci. Robot.* **6**, eaay2627 (2021).
5. F. Zhai, Y. Feng, Z. Li, Y. Xie, J. Ge, H. Wang, W. Qiu, W. Feng, 4D-printed untethered self-propelling soft robot with tactile perception: Rolling, racing, and exploring. *Matter.* **4**, 3313–3326 (2021).
6. J. Zhang, Y. Guo, W. Hu, M. Sitti, Wirelessly actuated thermo-and magneto-responsive soft bimorph materials with programmable shape-morphing. *Adv. Mater.* **33**, 2100336 (2021).
7. H. Shahsavan, A. Aghakhani, H. Zeng, Y. Guo, Z. S. Davidson, A. Priimagi, M. Sitti, Bioinspired underwater locomotion of light-driven liquid crystal gels. *Proc. Natl. Acad. Sci.* **117**, 5125–5133 (2020).
8. M. P. Da Cunha, M. G. Debije, A. P. Schenning, Bioinspired light-driven soft robots based on liquid crystal polymers. *Chem. Soc. Rev.* **49**, 6568–6578 (2020).
9. X. Ji, X. Liu, V. Cacucciolo, M. Imboden, Y. Civet, A. El Haitami, S. Cantin, Y. Perriard, H. Shea, An autonomous untethered fast soft robotic insect driven by low-voltage dielectric elastomer actuators. *Sci. Robot.* **4**, eaaz6451 (2019).
10. G. Li, X. Chen, F. Zhou, Y. Liang, Y. Xiao, X. Cao, Z. Zhang, M. Zhang, B. Wu, S. Yin, Self-powered soft robot in the Mariana Trench. *Nature.* **591**, 66–71 (2021).
11. W. Hu, G. Z. Lum, M. Mastrangeli, M. Sitti, Small-scale soft-bodied robot with multimodal locomotion. *Nature.* **554**, 81–85 (2018).
12. Y. Kim, H. Yuk, R. Zhao, S. A. Chester, X. Zhao, Printing ferromagnetic domains for untethered fast-transforming soft materials. *Nature.* **558**, 274–279 (2018).
13. Q. Cao, Q. Fan, Q. Chen, C. Liu, X. Han, L. Li, Recent advances in manipulating micro-and nano-objects with magnetic fields at small scales. *Mater. Horiz.* **7**, 638–666 (2020).
14. Y. Kim, X. Zhao, Magnetic soft materials and robots. *Chem. Rev.* **122**, 5317–5364 (2022).

15. J. Cui, T.-Y. Huang, Z. Luo, P. Testa, H. Gu, X.-Z. Chen, B. J. Nelson, L. J. Heyderman, Nanomagnetic encoding of shape-morphing micromachines. *Nature*. **575**, 164–168 (2019).

16. T. Xu, J. Zhang, M. Salehizadeh, O. Onaizah, E. Diller, Millimeter-scale flexible robots with programmable three-dimensional magnetization and motions. *Sci. Robot.* **4**, eaav4494 (2019).

17. Y. Alapan, A. C. Karacakol, S. N. Guzelhan, I. Isik, M. Sitti, Reprogrammable shape morphing of magnetic soft machines. *Sci. Adv.* **6**, eabc6414 (2020).

18. H. Deng, K. Sattari, Y. Xie, P. Liao, Z. Yan, J. Lin, Laser reprogramming magnetic anisotropy in soft composites for reconfigurable 3D shaping. *Nat. Commun.* **11**, 1–10 (2020).

19. Y. Ju, R. Hu, Y. Xie, J. Yao, X. Li, Y. Lv, X. Han, Q. Cao, L. Li, Reconfigurable magnetic soft robots with multimodal locomotion. *Nano Energy*. **87**, 106169 (2021).

20. L. Manamanchaiyaporn, T. Xu, X. Wu, Magnetic soft robot with the triangular head–tail morphology inspired by lateral undulation. *IEEE/ASME Trans. Mechatron.* **25**, 2688–2699 (2020).

21. N. Ebrahimi, C. Bi, D. J. Cappelleri, G. Ciuti, A. T. Conn, D. Faivre, N. Habibi, A. Hošovskỳ, V. Iacovacci, I. S. Khalil, Magnetic actuation methods in bio/soft robotics. *Adv. Funct. Mater.* **31**, 2005137 (2021).

22. Z. Ren, R. Zhang, R. H. Soon, Z. Liu, W. Hu, P. R. Onck, M. Sitti, Soft-bodied adaptive multimodal locomotion strategies in fluid-filled confined spaces. *Sci. Adv.* **7**, eabh2022.

23. L. Cao, D. Yu, Z. Xia, H. Wan, C. Liu, T. Yin, Z. He, Ferromagnetic Liquid Metal Putty-Like Material with Transformed Shape and Reconfigurable Polarity. *Adv. Mater.* **32**, 2000827 (2020).

24. Q. Ze, X. Kuang, S. Wu, J. Wong, S. M. Montgomery, R. Zhang, J. M. Kovitz, F. Yang, H. J. Qi, R. Zhao, Magnetic shape memory polymers with integrated multifunctional shape manipulation. *Adv. Mater.* **32**, 1906657 (2020).

25. X. Yang, W. Shang, H. Lu, Y. Liu, L. Yang, R. Tan, X. Wu, Y. Shen, An agglutinate magnetic spray transforms inanimate objects into millirobots for biomedical applications. *Sci. Robot.* **5**, eabc8191 (2020).

26. Y. Kim, G. A. Parada, S. Liu, X. Zhao, Ferromagnetic soft continuum robots. *Sci. Robot.* **4**, eaax7329 (2019).

27. Z. Ren, W. Hu, X. Dong, M. Sitti, Multi-functional soft-bodied jellyfish-like swimming. *Nat. Commun.* **10**, 2703 (2019).

28. Z. Zheng, H. Wang, L. Dong, Q. Shi, J. Li, T. Sun, Q. Huang, T. Fukuda, Ionic shape-morphing microrobotic end-effectors for environmentally adaptive targeting, releasing, and sampling. *Nat. Commun.* **12**, 1–12 (2021).

29. C. Zhou, Y. Yang, J. Wang, Q. Wu, Z. Gu, Y. Zhou, X. Liu, Y. Yang, H. Tang, Q. Ling, Ferromagnetic soft catheter robots for minimally invasive bioprinting. *Nat. Commun.* **12**, 1–12 (2021).

30. S. Wu, W. Hu, Q. Ze, M. Sitti, R. Zhao, Multifunctional magnetic soft composites: a review. *Multifunct. Mater.* **3**, 042003 (2020).

31. H.-J. Chung, A. M. Parsons, L. Zheng, Magnetically controlled soft robotics utilizing elastomers and gels in actuation: A review. *Adv. Intell. Syst.* **3**, 2000186 (2021).

32. E. Yarali, M. Baniasadi, A. Zolfagharian, M. Chavoshi, F. Arefi, M. Hossain, A. Bastola, M. Ansari, A. Foyouzat, A. Dabbagh, Magneto-/electro-responsive polymers toward manufacturing, characterization, and biomedical/soft robotic applications. *Appl. Mater. Today*. **26**, 101306 (2022).

33. P. R. Wurman, S. Barrett, K. Kawamoto, J. MacGlashan, K. Subramanian, T. J. Walsh, R. Capobianco, A. Devlic, F. Eckert, F. Fuchs, L. Gilpin, P. Khandelwal, V. Kompella, H. Lin, P. MacAlpine, D. Oller, T. Seno, C. Sherstan, M. D. Thomure, H. Aghabozorgi, L. Barrett, R. Douglas, D. Whitehead, P. Dürr, P. Stone, M. Spranger, H. Kitano, Outracing champion Gran Turismo drivers with deep reinforcement learning. *Nature*. **602**, 223–228 (2022).

34. V. Mnih, K. Kavukcuoglu, D. Silver, A. A. Rusu, J. Veness, M. G. Bellemare, A. Graves, M. Riedmiller, A. K. Fidjeland, G. Ostrovski, S. Petersen, C. Beattie, A. Sadik, I. Antonoglou, H. King, D. Kumaran, D. Wierstra, S. Legg, D. Hassabis, Human-level control through deep reinforcement learning. *Nature*. **518**, 529–533 (2015).

35. J. Degrave, F. Felici, J. Buchli, M. Neunert, B. Tracey, F. Carpanese, T. Ewalds, R. Hafner, A. Abdolmaleki, D. de las Casas, C. Donner, L. Fritz, C. Galperti, A. Huber, J. Keeling, M. Tsimpoukelli, J. Kay, A. Merle, J.-M. Moret, S. Noury, F. Pesamosca, D. Pfau, O. Sauter, C. Sommariva, S. Coda, B. Duval, A. Fasoli, P. Kohli, K. Kavukcuoglu, D. Hassabis, M. Riedmiller, Magnetic control of tokamak plasmas through deep reinforcement learning. *Nature*. **602**, 414–419 (2022).

36. J. Lee, J. Hwangbo, L. Wellhausen, V. Koltun, M. Hutter, Learning quadrupedal locomotion over challenging terrain. *Sci. Robot.* **5**, eabc5986 (2020).

37. S. Fujimoto, H. Hoof, D. Meger, paper presented at the 35th International Conference on Machine Learning, Stockholm, Sweden, July 2018.

38. R. Zhao, Y. Kim, S. A. Chester, P. Sharma, X. Zhao, Mechanics of hard-magnetic soft materials. *J. Mech. Phys. Solids*. **124**, 244–263 (2019).

39. W. Huang, X. Huang, C. Majidi, M. K. Jawed, Dynamic simulation of articulated soft robots. *Nat. Commun.* **11**, 2233 (2020).

40. M. Bergou, M. Wardetzky, S. Robinson, B. Audoly, E. Grinspun, Discrete elastic rods. *ACM Trans. Graph.* **27**, 1–12 (2008).

41. M. K. Jawed, F. Da, J. Joo, E. Grinspun, P. M. Reis, Coiling of elastic rods on rigid substrates. *Proc. Natl. Acad. Sci.* **111**, 14663–14668 (2014).

42. M. Gazzola, L. H. Dudte, A. G. McCormick, L. Mahadevan, Forward and inverse problems in the mechanics of soft filaments. *R. Soc. Open Sci.* **5**, 171628 (2018).

43. X. Zhang, F. K. Chan, T. Parthasarathy, M. Gazzola, Modeling and simulation of complex dynamic musculoskeletal architectures. *Nat. Commun.* **10**, 4825 (2019).

44. J. Till, V. Aloi, C. Rucker, Real-time dynamics of soft and continuum robots based on Cosserat rod models. *Int. J. Robot. Res.* **38**, 723–746 (2019).

45. X. Zhang, N. Naughton, T. Parthasarathy, M. Gazzola, Friction modulation in limbless, three-dimensional gaits and heterogeneous terrains. *Nat. Commun.* **12**, 6076 (2021).

46. N. Naughton, J. Sun, A. Tekinalp, T. Parthasarathy, G. Chowdhary, M. Gazzola, Elastica: a compliant mechanics environment for soft robotic control. *IEEE Robot. Autom. Lett.* **6**, 3389–3396 (2021).

47. D. Q. Cao, R. W. Tucker, Nonlinear dynamics of elastic rods using the Cosserat theory: Modelling and simulation. International Journal of Solids and Structures. **45**, 460–477 (2008).

48. R. D. Field, P. N. Anandakumaran, S. K. Sia, Soft medical microrobots: Design components and system integration. *Appl. Phys. Rev.* **6**, 041305 (2019).

49. E. Diller, J. Zhuang, G. Zhan Lum, M. R. Edwards, M. Sitti, Continuously distributed magnetization profile for millimeter-scale elastomeric undulatory swimming. *Appl. Phys. Lett.* **104**, 174101 (2014).

50. S. Wu, C. M. Hamel, Q. Ze, F. Yang, H. J. Qi, R. Zhao, Evolutionary algorithm-guided voxel-encoding printing of functional hard-magnetic soft active materials. *Adv. Intell. Syst.* **2**, 2000060 (2020).

51. S. Wu, Q. Ze, R. Zhang, N. Hu, Y. Cheng, F. Yang, R. Zhao, Symmetry-breaking actuation mechanism for soft robotics and active metamaterials. *ACS Appl. Mater. Interfaces*. **11**, 41649–41658 (2019).

52. X. Wang, G. Mao, J. Ge, M. Drack, G. S. Cañón Bermúdez, D. Wirthl, R. Illing, T. Kosub, L. Bischoff, C. Wang, J. Fassbender, M. Kaltenbrunner, D. Makarov, Untethered and ultrafast soft-bodied robots. *Commun. Mater.* **1**, 67 (2020).

53. R. S. Sutton, A. G. Barto, *Reinforcement learning: an introduction* (MIT Press, Cambridge, MA, 2018).

54. G. Brockman *et al.*, available at http://arxiv.org/abs/1606.01540.

55. T. Haarnoja, A. Zhou, P. Abbeel, S. Levine, paper presented at the 35th international conference on machine learning, Stockholm, Sweden, July 2018.

56. E. B. Joyee, Y. Pan, A fully three-dimensional printed inchworm-inspired soft robot with magnetic actuation. *Soft Robot.* **6**, 333–345 (2019).

57. H. Lu, Y. Hong, Y. Yang, Z. Yang, Y. Shen, Battery-less soft millirobot that can move, sense, and communicate remotely by coupling the magnetic and piezoelectric effects. *Adv. Sci.* **7**, 2000069 (2020).

58. S. O. Demir, U. Culha, A. C. Karacakol, A. Pena-Francesch, S. Trimpe, M. Sitti, Task space adaptation via the learning of gait controllers of magnetic soft millirobots. *Int. J. Robot. Res.* **40**, 1331–1351 (2021).

59. X. Li, Q. Cao, Z. Lai, S. Ouyang, N. Liu, M. Li, X. Han, L. Li, Bulging behavior of metallic tubes during the electromagnetic forming process in the presence of a background magnetic field. *J. Mater. Process. Technol.* **276**, 116411 (2020).

60. C. J. C. H. Watkins, P. Dayan, Q-learning. *Mach. Learn.* **8**, 279–292 (1992).

61. D. P. Kingma, J. Ba, paper presented at the 3rd International Conference for Learning Representations, San Diego, CA, May 2015.

62. T. P. Lillicrap, J. J. Hunt, A. Pritzel, N. Heess, T. Erez, Y. Tassa, D. Silver, D. Wierstra, paper presented at the 4th International Conference on Learning Representations, San Juan, PUR, May, 2016.

63. J. M. Lucas, M. S. Saccucci, Exponentially weighted moving average control schemes: properties and enhancements. *Technometrics*. **32**, 1–12 (1990).

64. H. van Hasselt, A. Guez, D. Silver, paper presented at the 30th AAAI Conference on Artificial Intelligence, Phoenix, AZ, February 2016.

65. G. Z. Lum, Z. Ye, X. Dong, H. Marvi, O. Erin, W. Hu, M. Sitti, Shape-programmable magnetic soft matter. Proc Natl Acad Sci USA. 113, E6007–E6015 (2016).

**Acknowledgments**
**Funding:**

National Natural Science Foundation of China (51821005).

Young Elite Scientists Sponsorship Program by CAST (YESS, 2018QNRC001).

**Author contributions:**
Conceptualization: Q.C., J.Y., X.H.
Methodology: J.Y., Q.C., X.H.
Investigation: J.Y., Y.J., Y.S., R.L.
Formal analysis: J.Y.
Visualization: J.Y., Y.J.
Funding acquisition: Q.C., X.H., L.L.
Project administration: Q.C., X.H.
Supervision: X.H., L.L., Q.C.
Writing – original draft: J.Y., Q.C., J.W., Y.S., R.L.
Writing – review & editing: Q.C., J.Y., X.H.

**Competing interests:** The authors declare that they have no competing interests.

**Data and materials availability:** All data, code, and materials used in the analyses are present in the text or the supplementary materials.

# Supplementary Materials for
# Adaptive actuation of magnetic soft robots using deep reinforcement learning

Jianpeng Yao, Quanliang Cao*, Yuwei Ju, Yuxuan Sun, Ruiqi Liu, Xiaotao Han*, Liang Li

*Corresponding author. Email: quanliangcao@hust.edu.cn (Q.C.); xthan@hust.edu.cn (X.H.)

**This PDF file includes:**

Figs. S1 to S6
Table S1
Legends for movies S1 to S5

**Other Supplementary Materials for this manuscript include the following:**

Movies S1 to S5
Codes
(Codes and Movies are available at https://github.com/alantes/RL-for-MSRs)

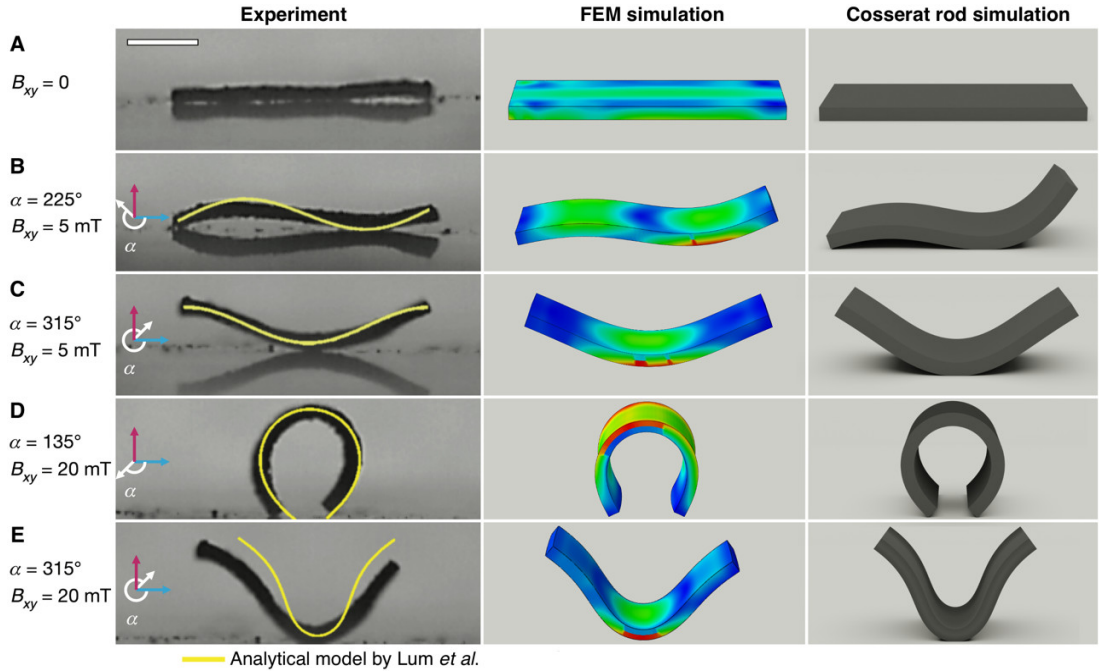

**Fig. S1. Static Cosserat rod simulation results and corresponding comparisons with experiments and FEM simulations.** The Experiments are from Hu *et al.* (*11*), and their analytical model is from Lum *et al.* (*65*), which is shown in yellow. The soft robot is of sinusoidal magnetization and its size is 3.7×1.5×0.185 mm. $\alpha$ is a clockwise angle from the *x*-axis, and it is used for describing the direction of the magnetic field. $B_{xy}$ is the amplitude of *xy* components of the field. (**A**) When the magnetic field is set to null, the soft robot and corresponding simulations have no deflection. (**B**-**E**) Cases where the magnetic field is not null. The deflection of the Cosserat rod simulations is slightly larger than experiments and FEM simulations, but both models correctly capture the deformation patterns under different magnetic fields.

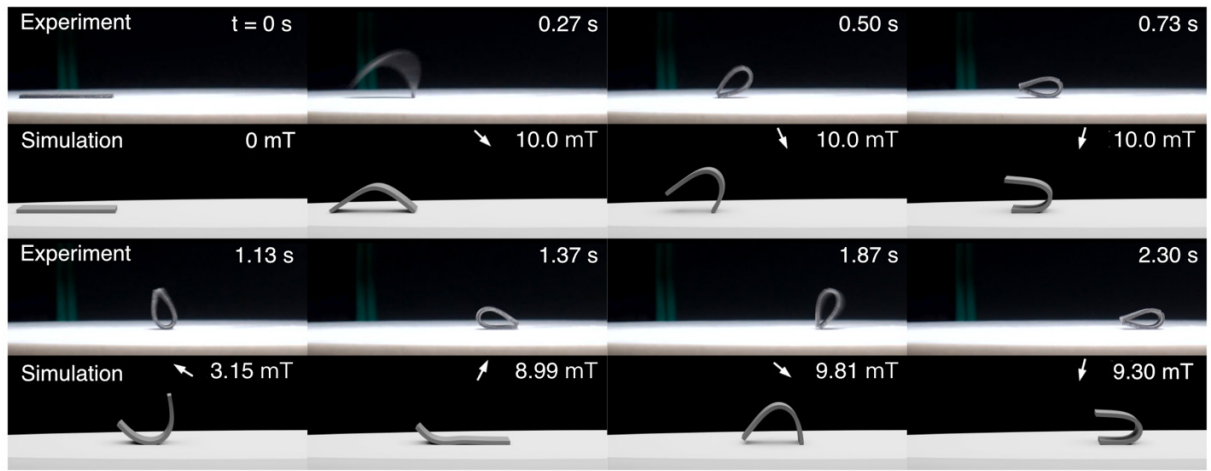

**Fig. S2. Comparisons between simulation and experiment of the robot with magnetization 2 under large field amplitude.** Two ends of the robot have opposite magnetization, making the ends easily attracted. Our model has not yet captured the influence of gradient magnetic force, so simulations and experiments do not correspond well. On the other hand, the structure of the robot itself is not stable and not a good choice for large deflection movement. See movie S5 for more.

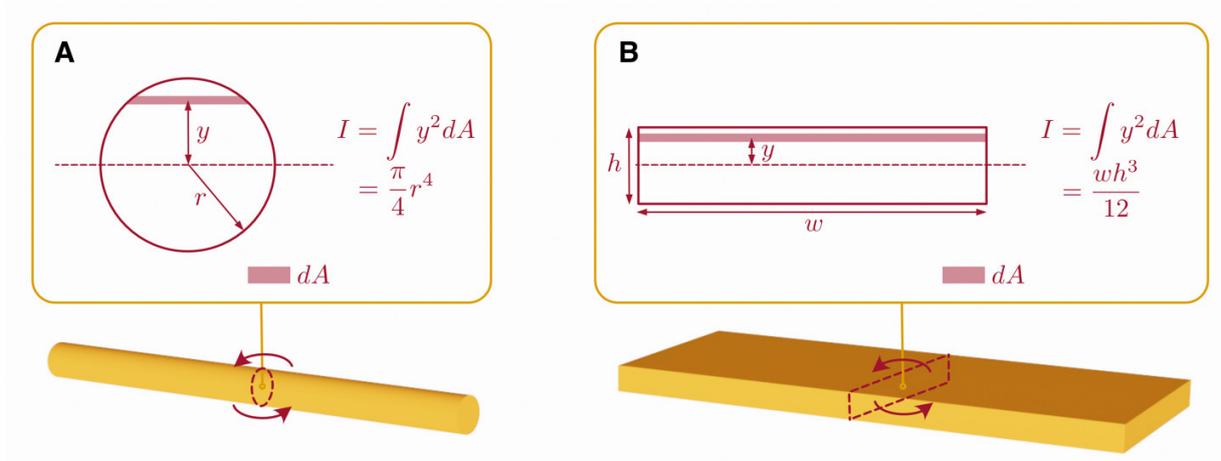

**Fig. S3. Rods with two types of cross sections and the methods of calculating the second moment of inertia.** Red curved arrows around the rods indicate bending about tangents. Dotted shapes in the middle of rods mean taking sections out for computing the second moment of inertia. (**A**) Calculating the second moment of inertia of a circular cross-section whose radius is $r$. The dotted line indicates the base axis for calculus. The area filled with light red means d$A$ in the equation. The calculated result is $\pi r^4/4$. (**B**) Calculating the second moment of inertia of a rectangular cross-section whose width is $w$ and height is $h$. The calculated result is $wh^3/12$.

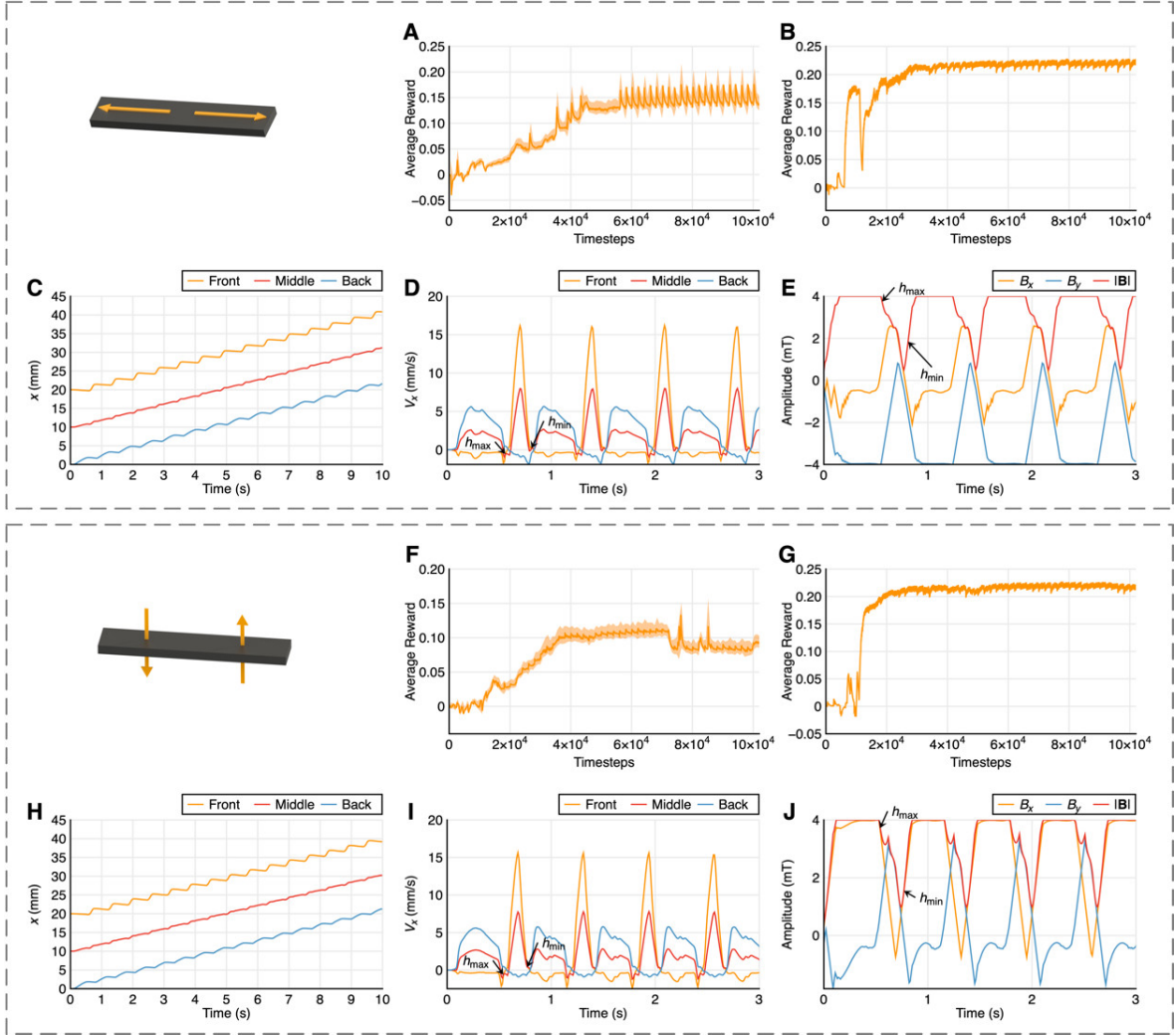

**Fig. S4. Learning results under small field amplitude.** (**A**) Average learning curve across 8 random seeds of the robot with magnetization pattern 1. In this and the following learning curves, rewards are smoothed using exponentially moving average (EMA). (**B**) Learning curve of the case we present of the robot with magnetization pattern 1. (**C**) Positions of the front, middle and back nodes in the case we present of the robot with magnetization pattern 1. (**D**) Velocities of the front, middle and back nodes in the case we present of the robot with magnetization pattern 1. I Field waveforms in the case we present of the robot with magnetization pattern 1. (**F-J**) Corresponding plots of the robot with magnetization pattern 2.

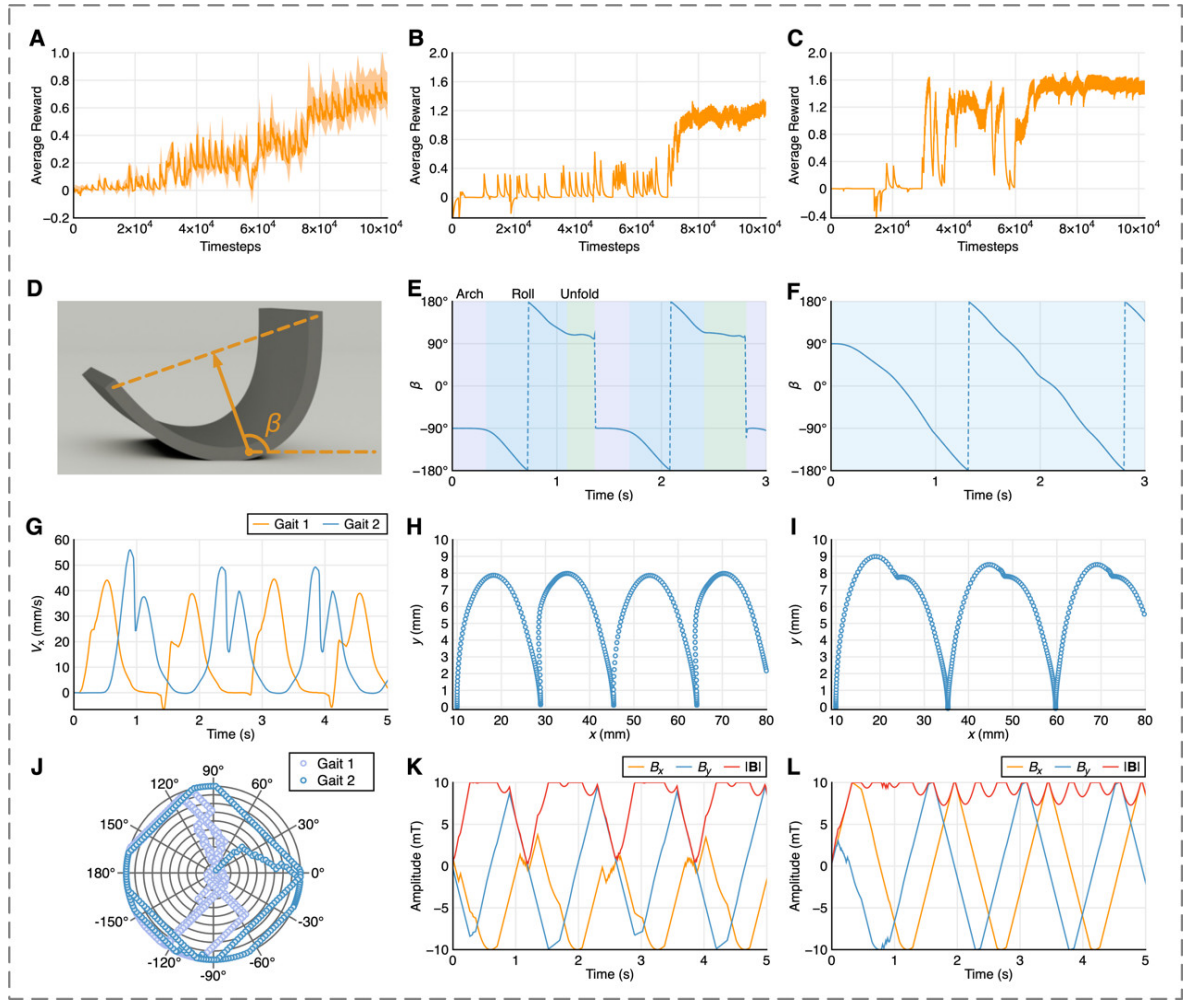

**Fig. S5. Learning results under large field amplitude of the robot with magnetization pattern 1.** (**A**) Average learning curve across 8 random seeds for the robot. (**B**) Learning curve of the case gait 1. (**C**) Learning curve of the case gait 2. (**D**) The definition of opening orientation angle $\beta$. (**E**) The change of orientation angle $\beta$ in gait 1. The motion of gait 1 can be separated into 3 phases. (**F**) The change of orientation angle $\beta$ in gait 2. The movement of gait 2 has only 1 phase. (**G**) Velocities of middle nodes in two gaits. (**H**) The trajectory of the middle node in gait 1. (**I**) Trajectory of the middle node in gait 2. (**J**) Magnetic field points in polar coordinate. (**K**) Field waveforms of gait 1. (**L**) Field waveforms of gait 1.

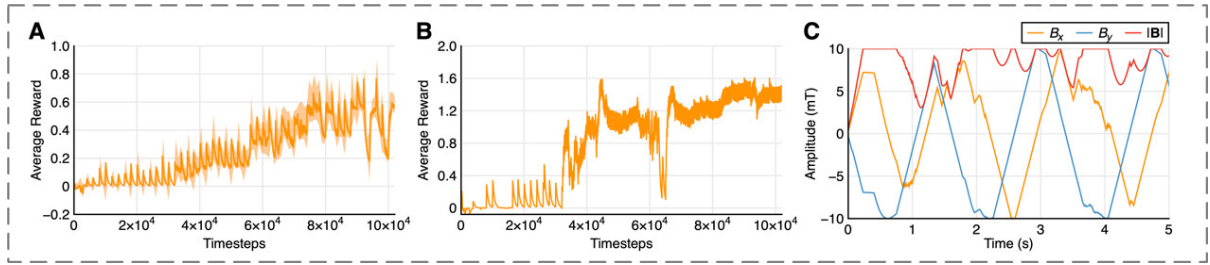

**Fig. S6. Learning results under large field amplitude of the robot with magnetization pattern 2.** (**A**) Average learning curve across 8 random seeds for the robot. (**B**) The learning curve of the case we present in our work. (**C**) Field waveforms in the case.

**Table S1. The three modes of deformation considered in our numerical model.** The directions of deformation are discussed in terms of tangent or normal vectors of cross-sections. $K$ stands for curvature, and $\sigma$ stands for strain. Subscripts t and n indicate directions of the quantities. t for tangent and n for normal. $B$ and $S$ correspond to rigidities. $E$ is Young's modulus. $I$ is the second moment of inertia. $G$ is the shear modulus. $A$ is the area of cross sections. The constant $\alpha_c$ equals to 4/3 for circular sections and equals to 1 for rectangular sections. $T$ stands for elastic torque generated due to bending, and $n$ stands for the elastic force generated due to shear or stretch. The superscript 0 indicates pre-strains. See (*42*) for more.

| Deformation modes | strains | rigidities | loads |
|---|---|---|---|
| Bending about tangent | $\kappa_t$ | $B_t = EI_t$ | $\tau_t = B_t(\kappa_t - \kappa_t^0)$ |
| Shear along tangent | $\sigma_t$ | $S_t = \alpha_c GA$ | $n_t = S_t(\sigma_t - \sigma_t^0)$ |
| Stretch along normal | $\sigma_n$ | $S_n = EA$ | $n_n = S_n(\sigma_n - \sigma_n^0)$ |

**Supplementary Movies**

**Movie S1.** The comparisons between simulations and experiments of the robot with magnetization 1 under relatively small field amplitude.

**Movie S2.** The comparisons between simulations and experiments of the robot with magnetization 2 under relatively small field amplitude.

**Movie S3.** The comparisons between simulations and experiments of the robot with magnetization 1 under relatively large field amplitude. Gait 1 is shown.

**Movie S4.** The comparisons between simulations and experiments of the robot with magnetization 1 under relatively large field amplitude. Gait 2 is shown.

**Movie S5.** The comparisons between simulations and experiments of the robot with magnetization 2 under relatively large field amplitude.